\documentclass{article}
\usepackage{microtype}
\usepackage{graphicx}
\usepackage{subfigure}
\usepackage{booktabs} 
\usepackage{wrapfig}
\usepackage{blindtext}
\usepackage{floatrow}
\newfloatcommand{capbtabbox}{table}[][\FBwidth]
\usepackage{amsthm}
\usepackage{amsmath}
\usepackage{bbm}
\usepackage{multirow}
\usepackage{graphicx}
\usepackage{subfigure}
\usepackage{bm}
\usepackage{babel}
\usepackage{caption}
\usepackage{amssymb}
\usepackage{diagbox}
\usepackage{colortbl}
\usepackage{subcaption}

\usepackage{hyperref}

\usepackage[accepted]{icml_arxiv}

\usepackage{amsmath}
\usepackage{amssymb}
\usepackage{mathtools}
\usepackage{amsthm}
\usepackage{tcolorbox}
\usepackage{paralist}
\usepackage[capitalize, noabbrev]{cleveref}

\theoremstyle{plain}

\theoremstyle{definition}

\theoremstyle{remark}

\usepackage[textsize=tiny]{todonotes}

\icmltitlerunning{LaRA: Benchmarking LLMs with Long-Context and Retrieval-Augmented Generation}

\begin{document}
\twocolumn[ \icmltitle{LaRA: Benchmarking Retrieval-Augmented Generation \\ and Long-Context LLMs - No Silver Bullet for LC or RAG Routing}



\icmlsetsymbol{equal}{*}

\begin{icmlauthorlist}
\icmlauthor{Kuan Li}{hkust} \icmlauthor{Liwen Zhang}{ali}
\icmlauthor{Yong Jiang}{ali} \icmlauthor{Pengjun Xie}{ali}
\icmlauthor{Fei Huang}{ali} \icmlauthor{Shuai Wang}{hkust}
\icmlauthor{Minhao Cheng}{psu}
\end{icmlauthorlist}

\icmlaffiliation{hkust}{The Hong Kong University of Science and Technology}
\icmlaffiliation{ali}{Tongyi Lab, Alibaba Group}
\icmlaffiliation{psu}{Penn State University}

\icmlcorrespondingauthor{Yong Jiang}{yongjiang.jy@alibaba-inc.com} \icmlcorrespondingauthor{Minhao Cheng}{mmc7149@psu.edu}

\icmlkeywords{Machine Learning, ICML}

\vskip 0.3in ]


\printAffiliationsAndNotice{The project was done during Kuan Li's internship at Tongyi Lab, Alibaba Group.}  

\begin{abstract}
Effectively incorporating external knowledge into Large Language Models (LLMs) is crucial for enhancing their capabilities and addressing real-world needs. Retrieval-Augmented Generation (RAG) offers an effective method for achieving this by retrieving the most relevant fragments into LLMs. However, the advancements in context window size for LLMs offer an alternative approach, raising the question of whether RAG remains necessary for effectively handling external knowledge. Several existing studies provide inconclusive comparisons between RAG and long-context (LC) LLMs, largely due to limitations in the benchmark designs. In this paper, we present LaRA, a novel benchmark specifically designed to rigorously compare RAG and LC LLMs. LaRA encompasses 2326 test cases across four practical QA task categories and three types of naturally occurring long texts. Through systematic evaluation of seven open-source and four proprietary LLMs, we find that the optimal choice between RAG and LC depends on a complex interplay of factors, including the model's parameter size, long-text capabilities, context length, task type, and the characteristics of the retrieved chunks. Our findings provide actionable guidelines for practitioners to effectively leverage both RAG and LC approaches in developing and deploying LLM applications. Our code and dataset is provided at: \href{https://github.com/Alibaba-NLP/LaRA}{\textbf{https://github.com/Alibaba-NLP/LaRA}}.
\end{abstract}

\section{Introducion}
While large language models (LLMs) excel across various domains, the dynamic nature of information poses significant challenges to their ability to acquire new knowledge effectively.  Current studies reveal several limitations of LLMs, including high computational costs when processing long texts, a tendency to produce factual errors and hallucinations, difficulty adapting to specialized domains, and a propensity for generating overly generic responses~\cite{DBLP:journals/corr/abs-2312-10997}. To address these limitations, researchers have explored retrieval-augmented generation (RAG)~\cite{DBLP:conf/icml/GuuLTPC20, DBLP:conf/nips/LewisPPPKGKLYR020}. RAG enables LLMs to efficiently utilize external knowledge by retrieving the most relevant fragments from uploaded documents, knowledge bases, or websites. However, recent advancements in LLMs, such as GPT-4o~\cite{openai2024a}, Llama 3.2~\cite{meta2024a}, Claude 3.5~\cite{anthropic}, and Qwen 2.5~\cite{yang2024qwen2}, now support input lengths of up to 128k tokens, offering an alternative by directly feeding the full context of relevant information into the model. This raises questions about the continued necessity of RAG, which was initially crucial for handling long texts, since these models can now potentially access and process the necessary information directly. Therefore, it is essential to systematically compare the strengths and weaknesses of RAG and long-context (LC) LLMs.

\textls[-10]{Numerous studies have investigated the performance differences between RAG and providing LLMs with full long contexts. For example, ~\citeauthor{DBLP:conf/iclr/0008PWM0LSBSC24} find that RAG outperforms LC on several traditional QA datasets. More recently, ~\citeauthor{DBLP:conf/emnlp/Li00MB24} argued that LC consistently outperforms RAG in almost all settings. However, ~\citeauthor{DBLP:journals/corr/abs-2409-01666} subsequently conducted experiments demonstrating that RAG is not inherently weaker than LC. This lack of consensus likely stems from several flaws in the evaluation pipeline design of existing benchmarks, including issues with the corpus (e.g., excessively short text lengths, failed replacements), evaluation metrics, and impractical task design.}


\textls[-10]{To address these issues and facilitate a robust comparison between RAG and LC, we propose LaRA, a benchmark for evaluating \underline{L}ong-context LLMs competing \underline{a}gaint \underline{RA}G. In constructing LaRA, we adhere to the following criteria: (1) Context length should be maximized within the LLMs' input limits to avoid truncation that could obscure the true capabilities of the models. (2) The context should consist of naturally occurring long text, rather than artificially constructed examples, to reflect real-world usage scenarios. (3) Answering questions should require information from the provided context, ensuring the LLM cannot answer them based solely on its internal knowledge. (4) Questions should have definitive answers to facilitate an accurate evaluation using LLMs. (5) The questions should reflect practical queries that humans are likely to ask in real-world LLM usage scenarios. LaRA is designed to serve as a guidebook for designing effective RAG-or-LC planning systems, therefore, it is crucial that all included questions are meaningful, relevant, and reflect real-world information needs.}

\textls[-10]{Specifically, LaRA includes three of the most common types of long contexts—novels, academic papers, and financial statements—selected to represent diverse writing styles and information densities. LaRA features four types of question-answering tasks designed to assess essential capabilities for handling long contexts: Locating specific information, Comparing different parts of the text, Reasoning about the content, and Detecting hallucinations. The QA pairs are constructed using a combination of human annotation and LLM-assisted generation, where we iteratively refine seed examples and prompts until a predefined pass rate is achieved for the generated pairs. To ensure accurate and reliable evaluation, we employ GPT-4o as a judge to determine the correctness of predictions and further validate the results by computing Cohen's Kappa coefficient between LLM and human evaluations, confirming a high level of consistency.}
 
\textbf{\emph{Our extensive experiments on LaRA demonstrate that the choice between RAG and LC is not trivial}}, as it varies significantly depending on factors such as model size, query type, type of tasks, context length, context type, and number of retrieved chunks. If we can pinpoint the scenarios in which RAG outperforms LC, we can better design workflows to route each query through RAG or LC, thereby optimizing for both cost and performance, leading to more efficient and effective LLM applications. Our key findings are as follows:

\begin{compactitem}
    \item \textbf{Model Strength}: RAG provides more significant improvements for weaker models. Our analysis indicates a correlation between model strength and RAG's effectiveness: the weaker the model, the greater the improvement from RAG. For instance, with a 128k context length, RAG outperformed LC by 6.48\% and 38.12\% in accuracy on Llama-3.2-3B-Instruct and Mistral-Nemo-12B, respectively. For models with strong long-text capabilities, such as GPT-4o and Claude-3.5-sonnet, LC generally outperforms RAG, demonstrating the effectiveness of these models in directly processing extensive contexts.
    \item \textbf{Context Length}: RAG's advantages become more pronounced as context length increases. With a 32k context length, LC achieved an average accuracy 2.4\% higher than RAG across all models. However, with a 128k context length, this trend reversed, with RAG outperforming LC by 3.68\%.
    \item \textbf{Task Performance}: RAG demonstrates similar performance to LC in single-location tasks and offers a significant advantage in identifying hallucinations. In contrast, LC excels in reasoning tasks and comparison tasks.
\end{compactitem}

\section{Revisiting RAG vs. Long Context Benchmarks}
Desipte a lot of benchmarks has been used to compare RAG with feeding lanuage model with long context, there lacks a clear guidelines and conclusion on when and where the RAG will be a better choice than long context. \citeauthor{DBLP:conf/iclr/0008PWM0LSBSC24} and \citeauthor{DBLP:conf/acl/BaiLZL0HDLZHDTL24} draw opposing conclusions on whether RAG or LC performs better on traditional QA datasets. Recently, ~\citeauthor{DBLP:conf/emnlp/Li00MB24} argued that LC consistently outperforms RAG in almost all settings, and ~\citeauthor{DBLP:journals/corr/abs-2409-01666} subsequently claim RAG can defeat LC on the same benchmark. In this section, we conduct a detailed analysis on the existing benchmark and analysis, and find the key issues are stem from some significant flaws with their evaluation pipeline. For simplicity, our analysis is mainly limited to question answering tasks based on long contexts. 

\subsection{Issues with External Contexts}

\paragraph{Insufficent context lengths.} As LLM base models continue to evolve, the definition of ``long context" has also shifted, expanding from the early limit of 4k tokens to the now commonly supported 128k context length. Some early work utilize datasets such as Qasper (QASP)~\cite{DBLP:conf/naacl/DasigiLBCSG21}, NarrativeQA (NQA)~\cite{DBLP:journals/tacl/KociskySBDHMG18}, and QuALITY (QLTY)~\cite{DBLP:conf/naacl/PangPJNPCPMT0B22} to compare RAG and LC. For instance, ~\citeauthor{DBLP:conf/iclr/0008PWM0LSBSC24} conduct experiments on these datasets and find that RAG can strengthen large models, such as Llama2-70B and GPT-43B. Similarly, ~\citeauthor{DBLP:journals/corr/abs-2406-13121} combine these datasets to create a new benchmark for further evaluations. However, such datasets no longer align with the current definition of long context. For example, QASP and QLTY have average context lengths of only 4912 and 6592 tokens, respectively, which are far below the context length capabilities of modern LLMs. Moreover, RAG typically uses chunk sizes of 300–600 tokens, and with 5–20 retrieved chunks, the total context length in RAG becomes comparable to that of full-context input, reducing the distinction between the two approaches. 

\paragraph{Data Leakage.}
Since LLMs use more and more datasets in the training procedure, the problem of data leakage becomes more serious. At the same time, it is challenging to verify whether these early datasets were part of the training data for LLMs, potentially causing the models to memorize the answers. For example, although NarrativeQA has an average context length of 84,770 tokens, Gemini 1.5pro achieves 100\% accuracy on this dataset~\cite{DBLP:journals/corr/abs-2406-13121}, indicating that either the dataset itself or its contexts were likely included in the model's training process. 

\textls[-10]{\paragraph{Inappropriate Contexts Handling.} 
$\infty$-bench~\cite{DBLP:conf/acl/ZhangCHXCH0TW0024}, a recent benchmark widely used for comparing RAG and LC~\cite{DBLP:conf/emnlp/Li00MB24, DBLP:journals/corr/abs-2409-01666}, includes two tasks, En.QA and En.MC, with average context lengths of 192.6k and 184.4k tokens, respectively. However, their method of handling the excessive context length is problematic. When the context exceeds the LLM's maximum context length, the middle portion of the context is truncated. Given that most models have a limit of 128k tokens or even less, it is highly likely that the answer to a question is removed during truncation. In such cases, failure to answer reflects the truncation issue rather than the model's capabilities. To verify this, we split the excessively long cases in En.QA and En.MC into several parts, each within the context length limit. These parts are then paired with the query and fed into the LLM separately. The LLM is instructed to first determine whether the question is answerable; if not, it would decline to provide an answer. After obtaining multiple answers, a voting mechanism is used to get the final answer. The results are list in Table~\ref{tab:vote}. We observe that after adopting the segmented input and voting mechanism, Qwen-2.5-7B can even outperform GPT-4o.}

\begin{table}[t]
\caption{Results on En.MC, EN.QA, and Zh.QA in $\infty$-bench.}
\resizebox{1\textwidth}{!}{\begin{tabular}{lccc}
\toprule Task  & GPT-4o& Qwen 2.5-7B & Qwen 2.5-7B (vote) \\
\midrule En.MC & 76.0  & 67.7& \textbf{87.9} \\
En.QA  & 31.5  & 20.3& \textbf{31.2} \\
En.QA (LLM)& 82.9  & 78.1& \textbf{85.5} \\
Zh.QA  & \textbf{26.3} & 18.4& 25.6  \\
Zh.QA (LLM)& 79.9  & 72.5& \textbf{85.2} \\
\bottomrule
\end{tabular}} 
\label{tab:vote}
\vspace{-0.5cm}
\end{table}

Moreover, to prevent overlap with data seen during LLM training, key entity replacement is employed as a countermeasure in $\infty$-bench. However, upon closer inspection, we find that some replacements are unsuccessful. For example, some entities mentioned in the questions  do not exist in the provided context and vice versa\footnote{\url{https://github.com/OpenBMB/InfiniteBench/issues/26}}.

\subsection{Inaccurate Evaluation}
\paragraph{Unreasonable metrics.}
Many previous evaluations use automated metrics such as F1-score and exact match (EM)~\cite{DBLP:conf/aaai/0011LH024, DBLP:conf/acl/ZhangFC24}, which are not reliable for NLG~\cite{DBLP:conf/emnlp/NovikovaDCR17}. For example, if the ground truth answer is ``Allyson Kalia" and the model's response is ``Allyson Kalia is convicted of the murder of Kiran's younger brother, Rosetta." the prediction is clearly correct. However, it would only achieve an F1-score of 0.29. This is also why the scores on the En.QA task in $\infty$-bench~\cite{DBLP:conf/acl/ZhangCHXCH0TW0024} tend to be very low. We use LLM to re-evaluate, and the results are shown in Table~\ref{tab:vote}. The accuracy becomes much higher, indicating that these datasets are not as difficult as they appear.

\textls[-15]{\paragraph{No dedicated benchmarks.}
Over the past year, several long-context benchmarks have been introduced, e.g., ZeroSCROLLS~\cite{DBLP:conf/emnlp/0002IEBL23}, LongBench~\cite{DBLP:conf/acl/BaiLZL0HDLZHDTL24}, BAMBOO~\cite{DBLP:conf/coling/DongTLZW24}, LooGLE~\cite{DBLP:conf/acl/LiWZZ24}, Ruler~\cite{DBLP:journals/corr/abs-2404-06654}, $\infty$-bench~\cite{DBLP:conf/acl/ZhangCHXCH0TW0024}, and LongBench-V2~\cite{bai2024longbench}, with text lengths progressively increasing from 20k to 200k tokens. However, these benchmarks focus primarily on testing the ability of models to handle long texts. Although some of these benchmarks include experiments related to RAG, they lack a more systematic comparison between RAG and LC. 
Furthermore, existing RAG benchmarks typically utilize context lengths under 10k tokens~\cite{DBLP:conf/aaai/0011LH024, DBLP:conf/acl/ZhangFC24, DBLP:conf/naacl/Stolfo24, DBLP:journals/corr/abs-2401-17043, DBLP:conf/naacl/Saad-FalconKPZ24}, failing to adequately address the challenges of long contexts. Although concurrent work like LONG$^2$RAG\cite{qi2024long2rag} explores long-context RAG, its average text length remains smaller than LaRA, and its focus lies on evaluating long-form responses. Similarly, Loong~\cite{DBLP:conf/emnlp/WangCCL0WYXZLLY24} introduces tasks for comparing RAG and LC in multi-document QA but suffers from homogeneity in queries. For example, the clustering task only involves determining citation relationships between papers, and reasoning questions merely require providing citation chains. These queries, disconnected from specific content and applicable to any similar text, fail to capture the generalization capabilities of LLMs and RAG across diverse scenarios and context distributions.}

\section{LaRA}
In this section, we introduce the construction of LaRA and how it addresses the issues present in previous benchmarks, as mentioned in Section 3. The statistics of LaRA are provided in Appendix \ref{apx:stat}.

\subsection{Long Context Data Collection}
In our context selection process, we adhere to the following principles: (1) Timeliness: We select \textbf{recent} high-quality long contexts to prevent data leakage issues, ensuring that they are less likely to have been included in the LLM's training data. (2) Appropriate Length: Considering that mainstream commercial and open-weight models typically support context length of 32k and 128k, we choose contexts that are as close to these window sizes as possible without exceeding them. (3) Naturalness: The chosen contexts are naturally occurring long documents, rather than artificially constructed or assembled from unrelated short texts, to ensure the benchmark reflects the complexity and diversity of real-world use. (4) Authoritativeness: All contexts are considered reliable and credible sources of information due to expertise, reputation, and qualifications of the authors or institutions behind them.

To ensure a diverse range of contexts, we select novels\footnote{\url{https://www.gutenberg.org/}}, financial statements\footnote{\url{https://www.annualreports.com/}}, and academic papers\footnote{\url{https://arxiv.org/}} as the context. For novels, we choose the txt format of novelettes and novels to serve as the 32k and 128k contexts, respectively. Financial statements include the latest quarterly reports (32k) and annual reports (128k) from publicly listed companies in the United States for the year 2024. To create contexts of appropriate length for academic papers, we combine several papers published on arXiv in 2024 that are related through citations.

\paragraph{Entity Replacement.} To mitigate the risk of data leakage from novels, which are likely present in LLMs' training data, we perform entity replacement. Previous work has employed similar strategies~\cite{DBLP:conf/acl/ZhangCHXCH0TW0024, DBLP:journals/tkde/LiSHL22}, but we find that many entity replacements were incorrect or inconsistent, leading to inaccurate evaluations. To address this, we use GPT-4o to accurately identify and replace character entities as well as formulating questions targeting the replaced entities, ensuring consistency between the novel text and the questions. Details are provided in Appendix \ref{apx:ner}.


\subsection{Tasks in LaRA}
To comprehensively evaluate the capabilities of LC LLMs and RAG, LaRA includes four major task categories: location, reasoning, comparison, and hallucination detection, which are designed to assess distinct aspects of LLM performance, motivated by the need to assess both the strengths and weaknesses of RAG and LC in handling complex, real-world information needs. Below, we will introduce each task in detail and further elaborate on the motivation behind them. Examples of each task are provided in Appendix \ref{apx:case}.

\paragraph{Location.} The location task, the most fundamental task in LaRA, evaluates an LLM's ability to locate specific information within a long context. In this task, the answer resides in a single sentence or paragraph within a long context, and no additional reasoning or computation is required to formulate a correct response, such as identifying a character's name or a specific value mentioned in the text. It is worth noting that the location task differs from the ``Needle in a Haystack" problem~\cite{needle2023}, which focuses on verbatim matching. In contrast, the location task allows for paraphrasing, as long as the underlying meaning is preserved. This task is crucial for assessing an LLM's basic comprehension and information retrieval capabilities within a long context.



\paragraph{Reasoning.} The reasoning task in LaRA involve questions that require logical deduction, inference, or calculation based on the information provided in the long context. Instead of directly extracting answers from the text, these tasks demand a deeper understanding and processing of the information to derive the correct answer, such as inferring the relationship between two characters or calculating relevant data in financial statements. These tasks evaluate the ability of LC and RAG to handle complex questions, particularly in scenarios where the long context contains a significant amount of noise irrelevant to the question. The specific questions vary significantly depending on the type of context involved. Instead of explicitly defining sub-task types, we adopt different seed questions tailored to specific text types. These seed questions are used to generate similar QA pairs through in-context learning. For example, in financial statements, which contain a significant amount of statistical data, we focus on computational questions, and for novels, the questions involve reasoning about the plot or character traits.



\paragraph{Comparison.} The comparison task in LaRA evaluates the ability of RAG and LC to synthesize information from multiple parts of a long context, comparing their content or numerical values to arrive at the final answer. Crucially, the comparison task also involves manually designing different seed questions tailored to various text types. This approach ensures that the generated questions are not only relevant but also reflect the nuances and complexities of the specific context. For instance, in academic papers, the questions may focus on comparing different explanations of the same phenomenon, while in novels, they may compare changes in a character’s traits or appearance over time. This task is essential for assessing an LLM's ability to extract information from different parts.

\paragraph{Hallucination detection.} Hallucination, a common issue in LLMs, occurs when the model generates inaccurate or irrelevant information~\cite{DBLP:journals/corr/abs-2311-05232}. The hallucination detection task aims to test the model's ability to decline answering questions that are not mentioned in the given context. Although the questions appear to be answerable using the context, the required information is not actually mentioned in the text. Consequently, such questions have a uniform answer: \emph{``XXX is not mentioned in the provided context."}
The ability to refuse to answer is crucial in practical applications of RAG and LC, particularly in domains where accuracy and reliability are paramount, as users cannot always guarantee that their questions have answers within the provided context. For example, a user might pose a seemingly relevant question about a paper, and if the model hallucinates and generates an incorrect response, it could be highly detrimental. 

\subsection{Data Annotation}
\label{sec:annotation}

The annotation process for different tasks follows a similar framework, starting with the manual creation of seed questions and answers. We then utilize GPT-4o to generate new QA pairs through in-context learning. A subset of newly generated QAs is sampled for manual validation to ensure correctness and practicality. If the pass rate does not meet a predefined threshold, the seed QAs and prompts are refined, followed by re-generation and re-validation. We provide the annotation prompt in Appendix \ref{apx:annotation}.

Annotating long texts presents a unique challenge due to the inherent difficulty of long context processing. One effective approach to improve generation quality is to convert annotations for long texts into annotations for shorter texts. To achieve this, we employ various strategies tailored to different context types and tasks. Specifically, for location and reasoning tasks, we split the long context into multiple segments, each approximately 10k tokens in length, and input them individually into GPT-4o to generate QAs. This approach serves multiple purposes: First, it reduces the cognitive load on the annotator (GPT-4o here) and improves the focus and accuracy of the generated QA pairs. Second, it ensures that the answers are evenly distributed across the entire context, as we observe that providing the full context to the LLM often results in answers being concentrated at the beginning and end of the context. Third, it allows us to examine the relationship between answer accuracy and answer location, enabling us to investigate whether the LLM suffers from the ``lost in the middle" issue, where performance declines for information located in the middle sections of long documents~\cite{DBLP:journals/tacl/LiuLHPBPL24}. For the comparison task, we split the context into smaller segments and then sample two segments to generate comparison questions similar to the seed questions. Meanwhile, our segmentation strategies are tailored to the specific context type to preserve the inherent structure and coherence of the documents. For research papers, we separate concatenated papers to maintain the integrity of each individual paper. For novels and financial statements, we directly split the text into multiple segments based on token count.

\subsection{Evaluation}

\begin{table*}
[t]
\centering
\resizebox{\textwidth}{!}{
\begin{tabular}{p{4.0cm}|m{0.8cm}m{0.8cm}|m{0.8cm}m{0.8cm}|m{0.8cm}m{0.8cm}|m{0.8cm}m{0.8cm}|m{0.8cm}m{0.8cm}}
\toprule \textbf{Model} & \multicolumn{2}{c|}{\textbf{Location}} & \multicolumn{2}{c|}{\textbf{Reasoning}}& \multicolumn{2}{c|}{\textbf{Comparison}}& \multicolumn{2}{c|}{\textbf{Hallucination}} & \multicolumn{2}{c}{\textbf{Overall}}  \\
\midrule \multicolumn{11}{c}{\emph{Open-source LLMs (32k)}}  \\
Llama-3.1-8B-Instruct   & 77.60  & 73.85\cellcolor{gray!30}   & 51.65   & 46.59\cellcolor{gray!30}& 58.29& 38.06\cellcolor{gray!30}  & 78.80  & 83.73\cellcolor{gray!30} & 66.58  & 60.56\cellcolor{gray!30}  \\
Llama-3.2-3B-Instruct   & 69.11  & 71.08\cellcolor{gray!30}   & 36.60   & 34.88\cellcolor{gray!30}& 31.10& 25.02\cellcolor{gray!30}  & 65.13  & 79.71\cellcolor{gray!30} & 50.48  & 52.67\cellcolor{gray!30}  \\
Llama-3.3-70B-Instruct  & 77.50  & 79.83\cellcolor{gray!30}   & 66.43   & 57.23\cellcolor{gray!30}& 65.33& \textbf{53.13}\cellcolor{gray!30} & 73.59  & 86.62\cellcolor{gray!30} & 70.71  & 69.20\cellcolor{gray!30}  \\
Llama-3.3-70B-Instruct-Q8   & 78.24  & 77.86\cellcolor{gray!30}   & 61.12   & 58.55\cellcolor{gray!30}& 64.83& 47.90\cellcolor{gray!30}  & 74.09  & 83.95\cellcolor{gray!30} & 69.57  & 67.06\cellcolor{gray!30}  \\
Qwen-2.5-7B-Instruct& 78.10  & 73.70\cellcolor{gray!30}   & 51.30   & 48.63\cellcolor{gray!30}& 62.35& 43.79\cellcolor{gray!30}  & 76.42  & 84.36\cellcolor{gray!30} & 67.04  & 62.62\cellcolor{gray!30}  \\
Qwen-2.5-72B-Instruct   & \textbf{83.29} & \textbf{81.63}\cellcolor{gray!30}  & \textbf{74.06}  & \textbf{65.24}\cellcolor{gray!30}   & \textbf{68.89}   & 49.24\cellcolor{gray!30}  & \textbf{85.08} & 83.76\cellcolor{gray!30} & \textbf{77.83} & \textbf{69.97}\cellcolor{gray!30} \\
Mistral-Nemo-12B& 54.45  & 72.38\cellcolor{gray!30}   & 29.81   & 45.74\cellcolor{gray!30}& 29.02& 43.63\cellcolor{gray!30}  & 35.27  & 71.70\cellcolor{gray!30} & 37.14  & 58.36\cellcolor{gray!30}  \\
\multicolumn{11}{c}{\emph{Proprietary LLMs (32k)}}   \\
GPT-4o  & \textbf{86.33} & 82.55\cellcolor{gray!30}   & 75.48   & \textbf{68.51}\cellcolor{gray!30}   & \textbf{79.66}   & 46.56\cellcolor{gray!30}  & 72.30  & 78.39\cellcolor{gray!30} & \textbf{78.44} & 69.00\cellcolor{gray!30}  \\
GPT-4o-mini & 79.92  & 77.69 \cellcolor{gray!30}  & \textbf{77.65}  & 62.23\cellcolor{gray!30}& 73.91& 54.48\cellcolor{gray!30}  & 52.82  & 67.17\cellcolor{gray!30} & 71.08  & 65.39\cellcolor{gray!30}  \\
Claude-3.5-sonnet   & 85.82  & \textbf{83.10}\cellcolor{gray!30}  & 70.42   & 66.81\cellcolor{gray!30}& 66.28& 51.38\cellcolor{gray!30}  & \textbf{86.12} & 90.99\cellcolor{gray!30} & 77.16  & \textbf{73.07}\cellcolor{gray!30} \\
Gemini-1.5-pro  & 78.94  & 74.71\cellcolor{gray!30}   & 64.68   & 53.50\cellcolor{gray!30}& 77.34& \textbf{56.43}\cellcolor{gray!30} & 84.58  & 88.03\cellcolor{gray!30} & 76.39  & 68.17\cellcolor{gray!30}  \\
\rowcolor{pink!33}\textbf{Avg GAP}  & \multicolumn{2}{c|}{\textcolor{blue!70}{0.08}} & \multicolumn{2}{c|}{\textcolor{blue!70}{4.66}} & \multicolumn{2}{c|}{\textcolor{blue!70}{15.22}} & \multicolumn{2}{c|}{\textcolor{red!50}{-10.38}} & \multicolumn{2}{c}{\textcolor{blue!70}{2.40}} \\
\midrule 
\multicolumn{11}{c}{\emph{Open-source LLMs (128k)}} \\
Llama-3.1-8B-Instruct   & 72.64  & 72.65\cellcolor{gray!30}   & 48.48   & 47.73\cellcolor{gray!30}& 40.87& 22.27\cellcolor{gray!30}  & 58.40  & 78.36\cellcolor{gray!30} & 55.10  & 55.25\cellcolor{gray!30}  \\
Llama-3.2-3B-Instruct   & 60.96  & 68.96\cellcolor{gray!30}   & 33.99   & 38.86\cellcolor{gray!30}& 24.54& 20.20\cellcolor{gray!30}  & 52.06  & 69.45\cellcolor{gray!30} & 42.89  & 49.37\cellcolor{gray!30}  \\
Llama-3.3-70B-Instruct  & 74.54  & 78.11\cellcolor{gray!30}   & 52.98   & 59.05\cellcolor{gray!30}& 43.09& 26.69\cellcolor{gray!30}  & 44.00  & 77.67\cellcolor{gray!30} & 53.65  & 60.38\cellcolor{gray!30}  \\
Llama-3.3-70B-Instruct-Q8   & 72.44  & 77.99\cellcolor{gray!30}   & 53.97   & 58.32\cellcolor{gray!30}& 45.32& 29.61\cellcolor{gray!30}  & 38.18  & 76.87\cellcolor{gray!30} & 52.48  & \textbf{60.70}\cellcolor{gray!30} \\
Qwen-2.5-7B-Instruct& 68.94  & 74.08\cellcolor{gray!30}   & 44.69   & 50.93\cellcolor{gray!30}& 39.51& 29.17\cellcolor{gray!30}  & 42.52  & 71.02\cellcolor{gray!30} & 48.91  & 56.30\cellcolor{gray!30}  \\
Qwen-2.5-72B-Instruct   & \textbf{76.10} & \textbf{78.92}\cellcolor{gray!30}  & \textbf{65.25}  & \textbf{64.37}\cellcolor{gray!30}   & \textbf{54.62}   & \textbf{36.10}\cellcolor{gray!30} & \textbf{64.45} & 71.32\cellcolor{gray!30} & \textbf{65.11} & 62.68\cellcolor{gray!30}  \\
Mistral-Nemo-12B& 23.44  & 72.29\cellcolor{gray!30}   & 14.87   & 50.53\cellcolor{gray!30}& 7.77 & 32.59\cellcolor{gray!30}  & 14.77  & 57.92\cellcolor{gray!30} & 15.21  & 53.33\cellcolor{gray!30}  \\
\multicolumn{11}{c}{\emph{Proprietary LLMs (128k)}}  \\
GPT-4o  & \textbf{87.70} & \textbf{82.22}\cellcolor{gray!30}  & \textbf{79.35}  & \textbf{70.26}\cellcolor{gray!30}   & 64.74& 40.84\cellcolor{gray!30}  & 56.34  & 67.33\cellcolor{gray!30} & 72.03  & 65.16\cellcolor{gray!30}  \\
GPT-4o-mini & 79.86  & 80.43\cellcolor{gray!30}   & 67.80   & 65.15\cellcolor{gray!30}& \textbf{65.31}   & \textbf{41.29}\cellcolor{gray!30} & 32.08  & 59.31\cellcolor{gray!30} & 61.26  & 61.55\cellcolor{gray!30}  \\
Claude-3.5-sonnet   & 85.44  & 80.94\cellcolor{gray!30}   & 73.79   & 64.81\cellcolor{gray!30}& 58.35& 33.18\cellcolor{gray!30}  & \textbf{77.64} & 84.73\cellcolor{gray!30} & \textbf{73.81} & \textbf{65.92}\cellcolor{gray!30} \\
Gemini-1.5-pro  & 82.15  & 75.41\cellcolor{gray!30}   & 67.97   & 48.47\cellcolor{gray!30}& 61.67& 36.54\cellcolor{gray!30}  & 66.16  & 78.60\cellcolor{gray!30} & 69.49  & 59.75\cellcolor{gray!30}  \\
\rowcolor{pink!33}\textbf{Avg GAP}  & \multicolumn{2}{c|}{\textcolor{red!50}{-5.25}} & \multicolumn{2}{c|}{\textcolor{red!50}{-1.39}} & \multicolumn{2}{c|}{\textcolor{blue!70}{14.30}} & \multicolumn{2}{c|}{\textcolor{red!50}{-22.36}} & \multicolumn{2}{c}{\textcolor{red!50}{-3.68}} \\
\bottomrule
\end{tabular}
}
\caption{The accuracy of baselines evaluated by GPT-4o on LaRA (\%). We evaluate
performance for context lengths of 32k and 128k tokens separately, with
the gray background representing \colorbox{gray!30}{RAG} and the white
background representing LC. The highest-performing open-source and proprietary
LLMs for each task are highlighted in bold. ``\textbf{Avg GAP}" refers to
the difference between the average accuracy of LC and RAG across all models
for a specific task (calculated as LC minus RAG). Blue text indicates
that LC performs better, while red text indicates that RAG is better.}
\label{tab:main}

\end{table*}


\paragraph{Metrics.} Automated evaluation metrics, such as F1-score and ROUGE, can often produce lower scores, while LLM evaluations for QA tasks with definitive answers have demonstrated high precision~\cite{DBLP:conf/acl/ChiangL23, DBLP:conf/emnlp/LiuIXWXZ23}. Therefore, we provide GPT-4o with the query, the ground-truth answer, and the prediction, enabling it to assess the correctness of the response (details and prompt are provided in Appendix~\ref{apx:judge}). Since LaRA consists solely of QA pairs with clearly defined answers and no open-ended questions, using LLM as a judge is highly effective in ensuring accuracy and consistency in the evaluation process.

\paragraph{Manual Verification.} To ensure the quality and reliability of LaRA, we incorporate manual verification at two stages of the construction process. First, during the generation process, we employ a pipeline involving sampling, prompt refinement, and seed QA selection as manual adjustments. We find that the choice of seed questions has the most significant impact, possibly because LLMs perform much better in in-context learning compared to zero-shot question generation, demonstrating the importance of providing relevant examples for guiding the generation process (For details, see Appendix \ref{apx:annotation}). Second, we calculate the Cohen's Kappa coefficient between the evaluation from LLM and human to quantitatively assess the agreement between the LLM and human evaluations, ensuring consistency and reliability in the judgment process. We provide the details in Appendix~\ref{apx:judge}.

\section{Experiments}
\subsection{Experimental Settings}
\paragraph{Baselines.} To investigate the impact of various factors on RAG and LC performance, we evaluate a diverse set of 11 LLMs, encompassing both open-source and proprietary models. This includes seven open-source LLMs: Llama-3.2-3B-Instruct~\cite{meta2024a}, Llama-3.1-8B-Instruct~\cite{meta2024b}, Llama-3.3-70B-Instruct~\cite{meta2024c}, Llama-3.3-70B-Instruct-Q8~\cite{meta2024c} (utilizing FP8 quantization), Qwen-2.5-7B-Instruct~\cite{yang2024qwen2}, Qwen-2.5-72B-Instruct~\cite{yang2024qwen2}, and Mistral-Nemo-12B~\cite{mistral}. We also evaluate four advanced proprietary LLMs: GPT-4o~\cite{openai2024a}, GPT-4o-mini~\cite{openai2024a}, Claude-3.5-Sonnet~\cite{anthropic}, and Gemini-1.5-Pro~\cite{DBLP:journals/corr/abs-2403-05530}.

\paragraph{Implementation of RAG.} Our evaluation employs a standardized configuration with a chunk size of 600 tokens, 5 chunks per document, and an overlap of 100 tokens between chunks. We utilize GTE-large-en-v1.5~\cite{zhang2024mgte, li2023towards} for embedding extraction and employ a hybrid search strategy combining embedding similarity and BM25~\cite{robertson2009probabilistic}.

\subsection{Main Results and Analysis}
\paragraph{Overall performance.}
Table~\ref{tab:main} presents the evaluation results, comparing the performance of LC and RAG across various LLMs. The table uses a white background for LC results and a gray background for RAG results, with the highest-performing open-source and proprietary LLMs for each task highlighted in bold. Our analysis reveals a complex relationship between model architecture, context length, and performance. For open-source models at a 32k context length, LC generally outperforms RAG, with the exception of Llama-3.2-3B-Instruct and Mistral-Nemo-12B. However, this trend reverses at a 128k context length, where RAG demonstrates superior performance across most models.  In contrast, proprietary models consistently favor LC at both context lengths, likely due to their larger parameter sizes and enhanced ability to process long-context inputs. The inherent self-attention mechanism in these models appears more effective at handling extended contexts compared to the sparse attention employed in RAG.
Notably, at a 128k context length, the top three performing models (GPT-4o, Gemini-1.5-Pro, and Claude-3.5-Sonnet) all utilize LC, while the bottom three (Llama-3.2-3B-Instruct, Qwen-2.5-7B-Instruct, and Mistral-Nemo-12B) are also LC-based. This observation underscores the absence of a universal ``winner" between RAG and LC, as performance is highly dependent on the specific LLM and context length.

\paragraph{Scaling law holds in LC.}
Our experimental results confirm the established scaling law in LC~\cite{DBLP:journals/corr/abs-2001-08361}: larger models consistently outperform smaller counterparts. For example, GPT-4o and Qwen-2.5-72B-Instruct show significant performance gains ranging from 7.35\% to 16.2\% over their smaller versions, GPT-4o-mini and Qwen-2.5-7B-Instruct, respectively. This advantage is further amplified at a 128k context length. While all models experience a performance decline with longer contexts, the drop is more pronounced for smaller models, highlighting their limitations in processing extensive textual input. This observation challenges the purported ability of smaller models to handle extremely long contexts effectively. 

\paragraph{RAG empowers models to handle extremely long context.}
At a 128k context length, RAG consistently outperforms LC across nearly all open-source models. For example, Llama-3.1-8B-Instruct and Qwen-2.5-7B-Instruct demonstrate improvements of 0.15\% and 7.39\%, respectively, when using RAG instead of LC. All models exhibit a performance decline at a 128k context length compared to 32k. However, LC experiences a more significant drop than RAG, indicating that as context length approaches its limit, RAG is less affected by the increase in context length. Furthermore, RAG enables models with weaker long-context capabilities, such as Mistral-Nemo-13B and Llama-3.2-3B-Instruct, to achieve performance comparable to other models. These findings highlight that while larger models excel at long-context processing, RAG offers an effective alternative for smaller or weaker models, ensuring competitive performance even with extended context lengths.

\begin{figure}[t]
\centering
\includegraphics[width=\linewidth]{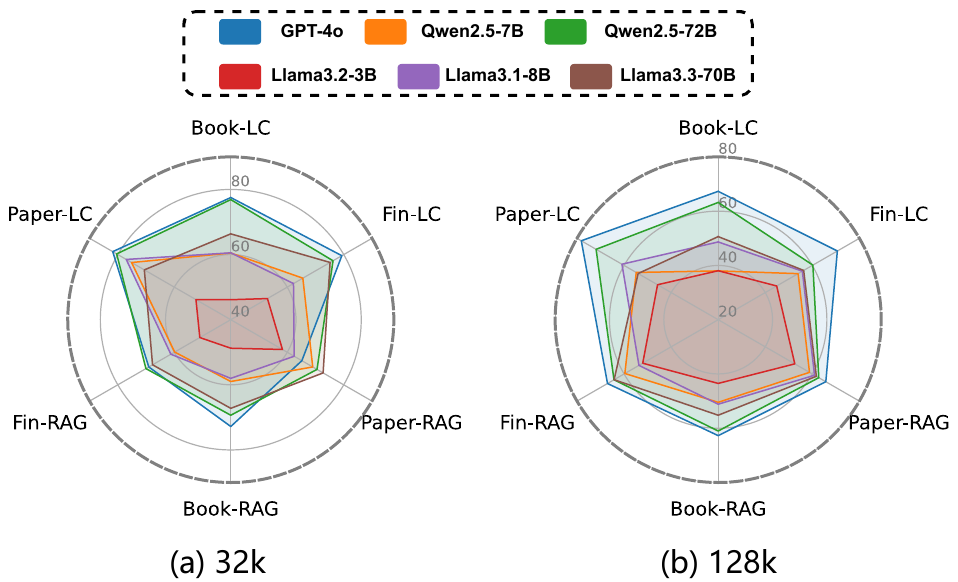}
\caption{The average accuracy across different context types. The left figure
(a) represents a context length of 32k, while the right figure (b)
represents a context length of 128k.}
\label{fig:context}

\vspace{-0.5cm}
\end{figure}

\subsection{Task Analysis}
\begin{figure*}[t]
  \centering
  \subfigure[Chunk quantity (32k)]{
  \begin{minipage}[t]{0.24\linewidth}
  \centering
  \includegraphics[width=1.04\linewidth]{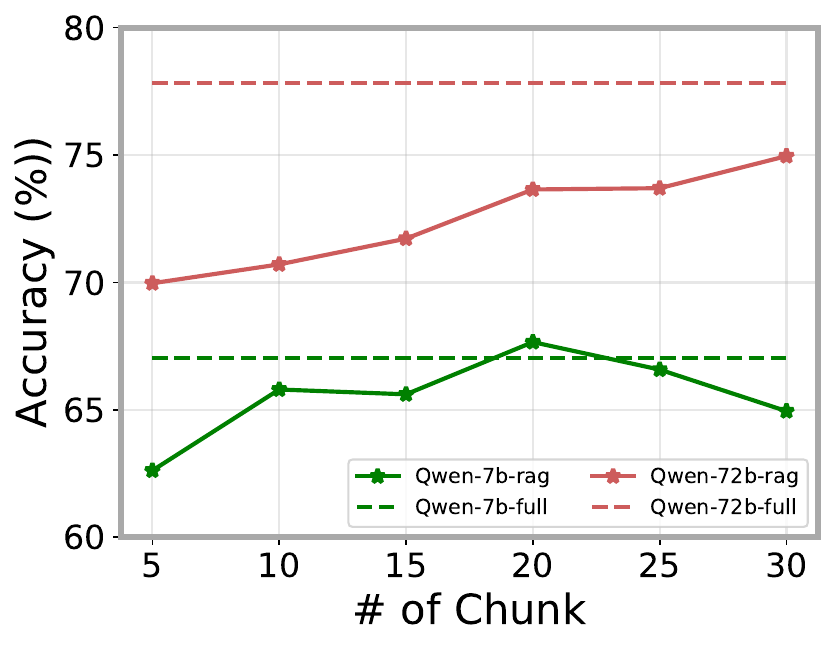}
  \end{minipage}
  }%
  \subfigure[Chunk quantity (128k)]{
  \begin{minipage}[t]{0.24\linewidth}
  \centering
  \includegraphics[width=\linewidth]{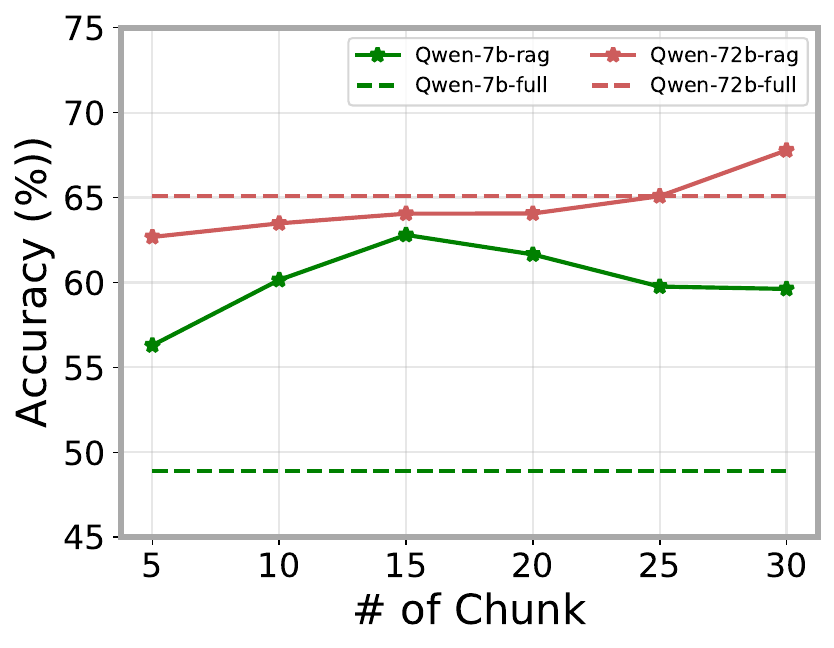}
  \end{minipage}
  }%
  \subfigure[Chunk size (32k)]{
  \begin{minipage}[t]{0.24\linewidth}
  \centering
  \includegraphics[width=1.04\linewidth]{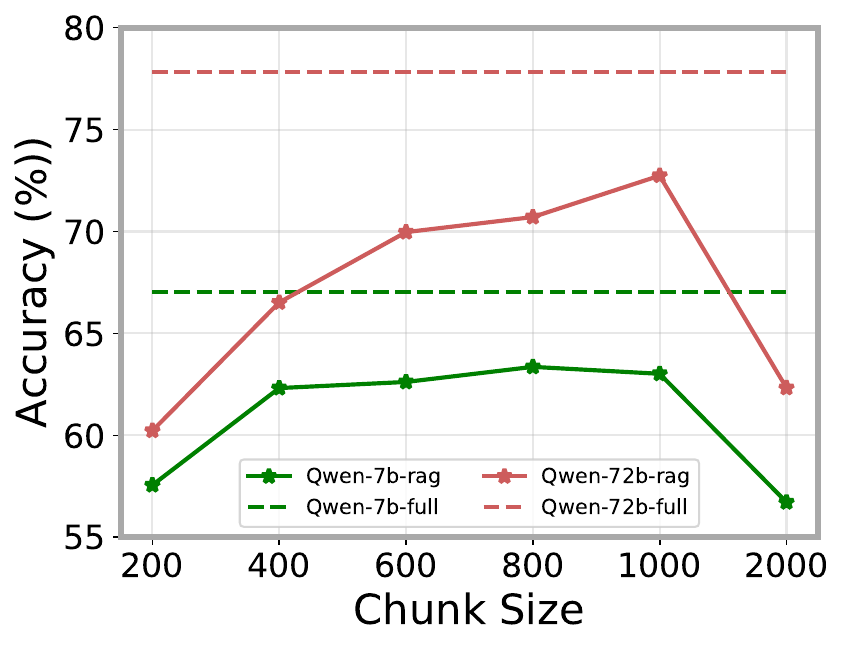}
  \end{minipage}
  }%
  \subfigure[Chunk size (128k)]{
  \begin{minipage}[t]{0.24\linewidth}
  \centering
  \includegraphics[width=1.04\linewidth]{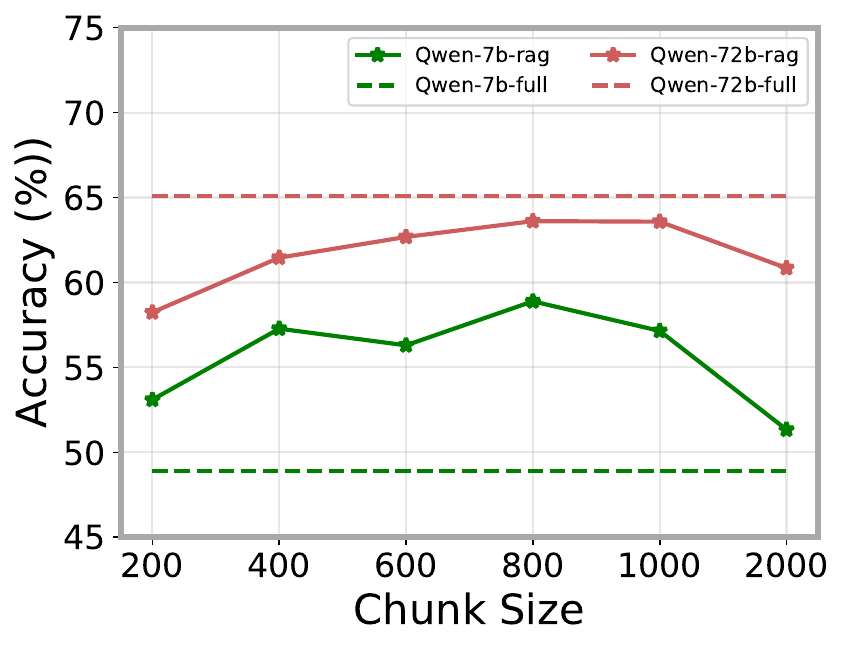}
  \end{minipage}
  }%
  \caption{The accuracy of Qwen-2.5-72B-Instruct and Qwen-2.5-7B-Instruct with different chunk quantity and size on LaRA.}
  \vspace{-0.6cm}
  \label{fig:chunk}
\end{figure*}

Table~\ref{tab:main} presents a detailed breakdown of LC and RAG performance across four distinct tasks. To provide a clear overview of the relative strengths of each task, we also calculate the average performance gap (avg gap) for each task, representing the mean difference in accuracy between LC and RAG across all evaluated models.
\textls[-10]{\paragraph{Location.}
The location task proves to be the easiest among the four, with both RAG and LC achieving high accuracy.  The average performance gap between RAG and LC is \textcolor{blue!70}{0.08\%} at a 32k context length and \textcolor{red!50}{-5.25\%} at 128k, indicating a slight advantage for LC with shorter contexts and a more pronounced advantage for RAG with longer contexts. For open-source models at 32k, the performance difference between RAG and LC is minimal, while at 128k, RAG demonstrates superior performance. This suggests that when models struggle to process long texts, retrieval acts as a valuable tool for location-based questions. Conversely, for proprietary models, LC consistently outperforms RAG, indicating that with sufficient model capacity, LLMs can outperform RAG in handling such simple tasks on their own.}

\paragraph{Reasoning.}
The performance trends for reasoning tasks basically mirror those observed in the location tasks, particularly for smaller models. With a 32k context, RAG exhibits slightly lower accuracy compared to LC, while at 128k, this trend reverses. However, for larger models like GPT-4o and Claude-3.5-Sonnet, the advantage of LC becomes more pronounced. At a 128k context length, GPT-4o and Claude-3.5-Sonnet outperform RAG by 9.09\% and 8.98\% in accuracy, respectively. We speculate that while reasoning tasks often rely on specific text segments for answers, other parts of the document may contain supplementary information that aids in inference. Models with stronger long-context capabilities are better equipped to leverage this global knowledge, leading to improved performance in reasoning tasks.

\paragraph{Comparison.}
Comparative tasks pose the greatest challenge for RAG, showing the largest performance gap compared to LC. The average gap reaches \textcolor{blue!70}{15.22\%} at 32k and \textcolor{blue!70}{14.30\%} at 128k. A deeper analysis of problematic cases reveals two primary reasons for the struggle of RAG. First, some comparative questions emphasize one aspect of the comparison while providing limited information about the other, making it difficult for RAG to retrieve both necessary chunks for a correct answer. Second, certain queries describe the comparison in abstract terms rather than concrete details, hindering the effectiveness of similarity-based retrieval. This abstraction makes it challenging for RAG to locate all the relevant chunks for comparison. In contrast to location tasks, which involve pinpointing a single piece of information, comparative tasks require the accurate retrieval and comparison of multiple distinct chunks, significantly increasing the complexity for RAG.

\textls[-15]{\paragraph{Hallucination detection.}
This is the only task where RAG demonstrates a clear advantage in both small and large models. LC tends to generate more hallucinated or incorrect answers, likely due to the increased noise introduced by feeding the entire text to the model. This makes the model more susceptible to errors and distractions, leading to fabricated responses. Interestingly, larger models do not exhibit an evident advantage in this task. While GPT-4o achieves top performance on other tasks, it only attains an accuracy of 56.34\% on hallucination detection. This suggests that even models capable of handling long contexts can be overwhelmed by the sheer volume of information and generate flawed conclusions. In contrast, RAG's selective retrieval of relevant information helps mitigate this issue, enabling both small and large models to better identify situations where they lack sufficient knowledge to provide an accurate answer.}

\subsection{Context Type Analysis}
\textls[-15]{We present the influence of different context types on the performance of RAG and LC in Figure~\ref{fig:context}. Novel-related questions pose the greatest challenge, while paper-related questions are the easiest. This disparity likely stems from the repetitive sentence structures common in novels, which can hinder precise information localization. Conversely, academic papers typically exhibit a stronger logical flow and higher information density, facilitating easier distinction between questions and answers. At a 32k context length, LC outperforms RAG for nearly all models. However, at 128k, weaker models demonstrate better performance with RAG. Interestingly, the performance gap between RAG and LC is smaller for novel-related tasks compared to paper-related or financial statements tasks, regardless of context length. This suggests that for less structured contexts, RAG presents a viable alternative for reducing computational cost. On the other hand, for highly structured texts like academic papers and financial statements, LC demonstrates a clear advantage.}

\subsection{Impact of Chunk Quantity and Size}
\textls[-10]{We explore the impact of the length of retrieved information length on RAG performance through two aspects: the number of chunks and the size of each chunk. We conduct experiments on Qwen-2.5-72B-Instruct and Qwen-2.5-7B-Instruct to observe the impact of chunk size and quantity on both large and small models. As shown in Figure~\ref{fig:chunk}, for the 72B model, performance improves consistently as the number of retrieved chunks increases, benefiting from its stronger long-context processing capability. In contrast, the 7B model exhibits a peak performance at an intermediate chunk quantity, after which excessive retrieval introduces noise that outweighs the information gain. Regarding chunk size, both excessively large and small chunks lead to performance degradation. Within a reasonable range, increasing chunk size provides some improvement, though its effect is less significant than increasing the number of chunks.}

\subsection{Lost in The Middle}
By controlling the location of the information required to answer the questions, we find that LC LLMs exhibit a decrease in accuracy when the answer is closer to the center of the context, indicating a susceptibility to the ``lost in the middle" phenomenon. In contrast, we do not observe a clear correlation between RAG performance and the position of the answer, suggesting that RAG models are more robust to this issue. These findings highlight a potential advantage of RAG models in handling long contexts, particularly for tasks that require accessing information from various parts of the document. For detailed results, see Appendix~\ref{apx:lost}.

\section{Conclusion}
This study addressed the critical question of whether RAG or LC is more effective for incorporating external knowledge into LLMs. Through the development and evaluation of LaRA, a novel benchmark, we demonstrated that the optimal choice depends on a complex interplay of factors, including model size, context length, and task type. Our findings challenge previous inconclusive comparisons and offer actionable guidelines for practitioners. LaRA serves as a valuable resource for evaluating and comparing RAG and LC models, facilitating further research in this rapidly evolving field.


\section*{Impact Statement}
This paper presents work whose goal is to advance the field of Machine Learning. There are many potential societal consequences of our work, none which we feel must be specifically highlighted here.


\bibliography{main}
\bibliographystyle{icml2025}

\newpage
\appendix
\onecolumn

\section{Statistics of LaRA}
\label{apx:stat}
LaRA consists of approximately 2300 test cases, encompassing three context types and four task categories. To ensure that the token count of the context is as close as possible to 32k and 128k without exceeding these limits, the primary token ranges for these two lengths are 20-30k and 80-120k, respectively. The average token counts and numbers of tasks are provided in Table~\ref{tab:length}.

\begin{table}[h]
    \centering
    \begin{tabular}{l|c|c|c|c}
    \toprule
    \textbf{Context} & \textbf{Location} & \textbf{Reasoning} & \textbf{Comparison} & \textbf{Hallucination} \\
    \midrule
    \rowcolor{gray!33}\multicolumn{5}{c}{\textbf{32k length}} \\
    \midrule
    Novel & 25673 & 25908 & 25681 & 25433 \\
    Financial & 27548 & 27531 & 27546 & 27527 \\
    Paper & 28078 & 28088 & 27708 & 28081 \\
    \textbf{\# of Cases} & 276 & 230 & 151 & 230 \\
    \midrule
    \rowcolor{gray!33}\multicolumn{5}{c}{\textbf{128k length}} \\
    \midrule
    Novel & 96452 & 96226 & 95903 & 96182 \\
    Financial & 92684 & 92831 & 92830 & 92812 \\
    Paper & 93911 & 93818 & 94731 & 93890 \\
    \textbf{\# of Cases} & 489 & 374 & 198 & 378\\
    \bottomrule
    \end{tabular}
    \caption{Statistics of LaRA.}
    \label{tab:length}
\end{table}

\section{Named Entity Recognition and Replacement for Novels}
\label{apx:ner}
We initially tried using some traditional Named Entity Recognition (NER) methods~\cite{wang-etal-2021-improving, wang-etal-2022-damo}, but found that their performance was poor. First, traditional sequence labeling models struggle with long-text processing due to fixed-length context windows, failing to capture cross-paragraph entity associations. Second, performance degrades significantly on out-of-distribution data, particularly when handling domain-specific or stylistically unique texts~\cite{DBLP:journals/tkde/LiSHL22}. These constraints prove especially problematic for literary analysis, where novels exhibit both long-range narrative dependencies and rich variations in entity references (e.g., honorifics, epithets, and contextual substitutions).

The emergence of LLM presents new opportunities to overcome these limitations through their superior contextual understanding and robust generalization capabilities. Hence, we leverage GPT-4o to perform entity extraction and replacement in full-length novels through a three-stage pipeline:

\begin{enumerate}
    \item We partition input texts into coherent segments averaging 500 tokens, preserving complete sentence/paragraph boundaries. This chunk size optimizes the balance between LLMs' context window constraints and narrative continuity requirements.
    \item Each text chunk undergoes parallel entity recognition through GPT-4o processing. Our extraction protocol captures both original names and contextual variants (e.g., ``The Dark Lord" $\rightarrow$ ``Voldemort" in Harry Potter). After feeding each chunk for extraction, we merge all the entities
    and remove duplicates to obtain a final list of entities while preserving legitimate aliases.
    \item For alias disambiguation, we prompt GPT-4o with the entity list and the novel's title to determine if multiple names indicate the same character, and if they do, we replace them with the same fake name.
\end{enumerate}

Figure~\ref{fig:prompt_ner} and Figure~\ref{fig:prompt_replace} detail our carefully engineered prompts for entity extraction and substitution respectively. Note that the novels we select are classical literary works, which, along with extensive related discussions, are included in the LLMs' training data. As a result, we find that simply providing the name of the novel is sufficient to accurately map multiple aliases to the same character.

\begin{figure}[h]
\begin{tcolorbox}
[title=Example prompt used for \textbf{\textit{Named Entity
Extraction}}] \textbf{Task:} Your task is to extract named entities(only
person) from the given paragraph. Response with a list of entities,
and strings in the list should be enclosed in double quote. Below is
an example, and you need to use a style consistent with the example.
\newline
\newline
\textbf{Example:}
\newline
\emph{$<$paragraph$>$}Filby became pensive. “Clearly,” the Time Traveller
proceeded, “any real body must have extension in four directions: it
must have Length, Breadth, Thickness, and—Duration. But through a natural
infirmity of the flesh, which I will explain to you in a moment, we incline
to overlook this fact. There are really four dimensions, three which
we call the three planes of Space, and a fourth, Time. There is, however,
a tendency to draw an unreal distinction between the former three
dimensions and the latter, because it happens that our consciousness
moves intermittently in one direction along the latter from the beginning
to the end of our lives.”

“That,” said a very young man, making spasmodic efforts to relight his
cigar over the lamp; “that . . . very clear indeed.”

“Now, it is very remarkable that this is so extensively overlooked,”
continued the Time Traveller, with a slight accession of cheerfulness.
“Really this is what is meant by the Fourth Dimension, though some people
who talk about the Fourth Dimension do not know they mean it. It is
only another way of looking at Time. There is no difference between Time
and any of the three dimensions of Space except that our consciousness
moves along it. But some foolish people have got hold of the wrong
side of that idea. You have all heard what they have to say about this
Fourth Dimension?”

“I have not,” said the Provincial Mayor.\emph{$<$/paragraph$>$}
\newline
\newline
\textbf{Entities:} [``Filby", ``Time Traveller", ``Provincial Mayor"]
\newline
\newline
\emph{$<$paragraph$>$}\{\textcolor{red}{chunk}\}\emph{$<$/paragraph$>$}
\newline
\newline
\textbf{Entities:}
\end{tcolorbox}
\caption{The prompt for extracting named entities from novel chunks. The
chunk that need to be extracted is highlighted in red text.}
\label{fig:prompt_ner}
\end{figure}

\begin{figure}[h]
\begin{tcolorbox}
[title=Example prompt used for \textbf{\textit{Name Replacement}}]
\textbf{Task:} I have extracted the names of all the characters in
the novel \{\textcolor{red}{novel}\}, as shown in the list below. Your task
is to replace the names in a novel with fictitious ones.
\newline
\newline
Firstly, You need to determine based on the novel's content whether
any of these names refer to the same character. If they refer to the
same person, they should be assigned the same fictitious name. You should
respond with a Python dict only, with the keys being the names from
the novel and the values being the new fictitious names. Strings in the
dict should be enclosed in double quote
\newline
\newline
\textbf{Name list:} \{\textcolor{red}{name\_list}\}
\end{tcolorbox}
\caption{The prompt for replacing the names with fictitious ones.}
\label{fig:prompt_replace}
\end{figure}

\section{LLM as A Judge}
\label{apx:judge}
Given the unreliability of rule-based evaluations and the high costs associated with human evaluation, the use of LLM for assessment has gained increasing popularity~\cite{DBLP:conf/coling/LiuYHZHWDSZ24, DBLP:conf/emnlp/WangCCL0WYXZLLY24}. In LaRA, we prompt GPT-4o to determine whether a model correctly answer a question, using the query, the ground-truth answer, and the model's prediction as inputs. The specific prompt is shown in Figure~\ref{fig:judge_prompt}.

\begin{figure}[h]
\begin{tcolorbox}
[title=Example prompt used for \textbf{\textit{Evaluation}}]
\textbf{Task:} You are a discriminator that judges whether the predictions to questions are correct.
\newline
\newline
I will provide you with a question and its Ground-truth answer, as well as an answer from an AI assistant. You need to judge whether the AI assistant's answer is correct based on the Ground-truth answer. If it is correct, you should only output True; if it is incorrect, only output False.
\newline
\newline
\textbf{[Query]} \{\textcolor{red}{query}\}
\newline
\newline
\textbf{[Ground-truth Answer]} \{\textcolor{red}{label}\}
\newline
\newline
\textbf{[AI Assistant's Answer]} \{\textcolor{red}{pred}\}
\newline
\newline
\textbf{Your judgment:}
\end{tcolorbox}
\caption{The prompt for evaluation.}
\label{fig:judge_prompt}
\end{figure}

To verify the consistency between GPT-4o evaluation and human evaluation, we compute the Cohen's Kappa coefficient, which is defined as:
\begin{equation}
\kappa=\frac{P_o-P_e}{1-P_e},
\end{equation}
where $P_o$ represents the proportion of agreement between the two evaluators on the positive and negative classes, and $P_e$ denotes the probability of agreement between evaluators under the assumption of independent and random classification.:
\begin{equation}
    P_o=\frac{TP + TN}{TP + FP +TN + FN}.
\end{equation}
\begin{equation}
    P_e=\frac{(TP+FN)(TP+FP)+(TN+FN)(TN+FP)}{(TP + FP +TN + FN)^2},
\end{equation}
where TP and TN refer to cases where both the human evaluator and the model agree that the answer is correct, while FP and FN refer to cases where only one of them judges the answer as correct.

A Cohen’s Kappa coefficient greater than zero indicates consistency in the evaluation, with values approaching 1 indicating stronger agreement. We sample 100 predictions from each task type at a 128k context length from GPT-4o (LC) and Qwen-2.5-7B (LC) for manual evaluation and then compute the Cohen’s Kappa coefficient. The results are provided in Table~\ref{tab:kappa}. The values for both models are close to 1, demonstrating that evaluations from GPT-4o are highly consistent with human evaluation, whether applied to large or small models.

\begin{table}[h]
    \centering
    \begin{tabular}{l|cccc}
        \toprule
        \textbf{Model} & \textbf{Location} & \textbf{Reasoning} & \textbf{Comparison} & \textbf{Hallucination} \\
        \midrule
         Qwen-2.5-7B  & 0.9737 & 0.9184 & 0.8955 & 0.9800\\
         GPT-4o & 0.9509 & 0.9254 & 0.9776 & 0.9604 \\
        \bottomrule
    \end{tabular}
    \caption{Cohen’s Kappa coefficient for GPT-4o (LC) and Qwen-2.5-7B (LC).}
    \label{tab:kappa}
\end{table}

\section{Lost in The Middle}
\label{apx:lost}
To assess the presence of ``lost in the middle" in our LaRA benchmark, we ensure a uniform distribution of answers across different positions within the context during the annotation of location and reasoning tasks. Specifically, for location tasks with 32k and 128k context lengths, we divide the context into 5 and 10 segments, respectively, and for reasoning tasks, we use 3 and 6 segments for 32k and 128k contexts, respectively.

\begin{figure*}[t]
\includegraphics[width=\linewidth]{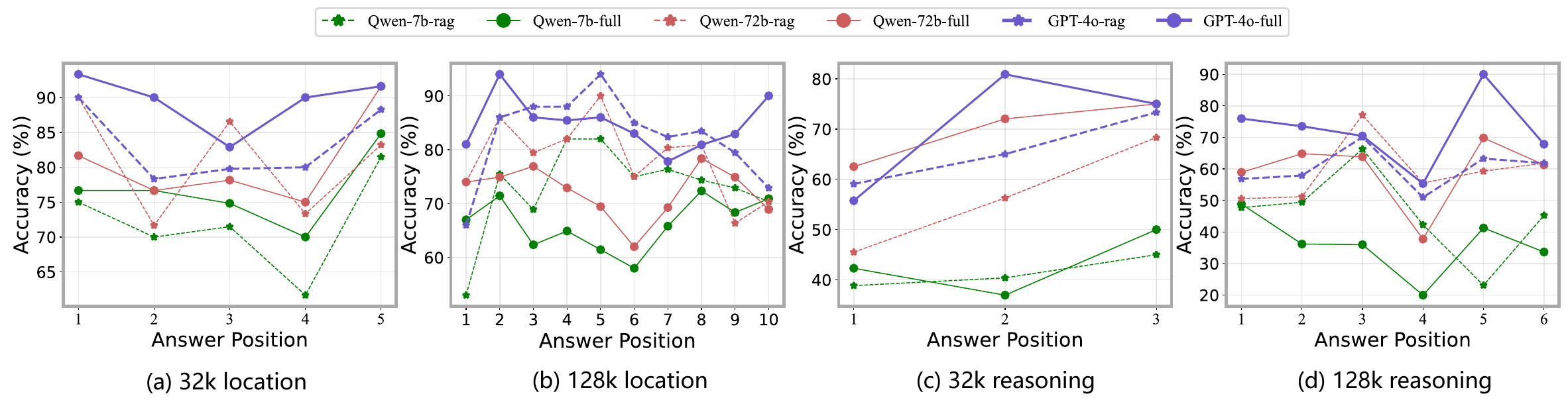}
\caption{The accuracy of the location and reasoning tasks when the
answer appears at different positions within the context. 32k-length and
128k-length contexts are split into 5 and 10 segments, respectively. The contexts here are novels and financial statements, as we split papers in another method metioned in Section \ref{sec:annotation}.}
\label{fig:lost}
\end{figure*}

As illustrated in Figure~\ref{fig:lost}, our experiments reveal that LC models indeed suffer from the ``lost in the middle" phenomenon. This issue is particularly pronounced in weaker models like Qwen-2.5-7B. In contrast, RAG demonstrates consistent accuracy regardless of answer position, showcasing its robustness against this phenomenon. This highlights a key advantage of RAG: its ability to effectively retrieve and utilize relevant information regardless of its location within the context.

\section{Annotation Details}
\label{apx:annotation}
Recent advancements in LLMs have demonstrated their ability to generate high-quality text comparable to human output, positioning synthetic data generated by LLMs as a viable alternative or complement to human-generated data~\cite{DBLP:conf/acl/LongWXZDCW24}. Benefiting from this, we use GPT-4o to generate highly diverse question-answer pairs across multiple scenarios and tasks.

Specifically, for each context type and task category, we design different prompts and seed questions based on distinct guiding principles. To better understand how we generate QA pairs in LaRA, we present the prompts used for generating financial statement QAs as an example. The prompts for the Location, Reasoning, Comparison, and Hallucination tasks are shown in Figures~\ref{fig:an_loc},~\ref{fig:an_rea}, ~\ref{fig:an_comp}, and ~\ref{fig:an_hallu}, respectively.

Each task’s annotations share some common principles, such as ensuring that questions are unanswerable without the provided context and keeping responses concise to avoid lengthy elaborations. Additionally, there are specific requirements to more precisely define the scope of the questions. For example, in the location task, we require the answer to be directly extracted from the context and prohibit any interpretive phrasing. In contrast, for the reasoning task, the answer must be impossible to obtain through simple lookup. Finally, seed example questions are provided for in-context learning, both as a reference and to ensure clear output formatting for easier subsequent processing. During fine-tuning prompts and seed questions, we find that once the problem types are precisely defined, the seed examples have a significant impact on the quality of the generated QA pairs. The higher the diversity of seed examples, the less uniform the generated questions become. Therefore, we strive to ensure that the seed questions cover various aspects and reflect the practical questions that users might realistically ask in real-world scenarios. For each task, we provide 3-5 seed examples.

\begin{figure}[h]
\begin{tcolorbox}
[title=Example prompt used for generating \textbf{\textit{Location}} QAs in financial statements]

\textbf{Task:} You are a financial question designer. Based on the provided financial statement, create a factual question and its corresponding answer that strictly satisfies these criteria:
\newline
\newline
\textbf{Design Requirements:}
\newline
\newline
1. Direct Extraction:

\qquad  - The answer must be explicitly stated in a single contiguous segment of the document

\qquad  - Require pinpoint localization (e.g., specific value, exact date, named section)
\newline
\newline
2. Answerability Constraints:

\qquad  - Unanswerable without the provided context

\qquad  - Must reference concrete elements (numerical values, named metrics, verbatim terms)
\newline
\newline
3. Response Specifications:

\qquad  - Maximum answer length: 20 words

\qquad  - Prohibit any interpretive phrasing
\newline
\newline
\textbf{Example QAs}:

\qquad  \{Q: ``What was the amount of impairment changes recorded to goodwill during the three months ended December 31, 2023?", A:``\$62.8 million."\}

\qquad  \{Q: ``How many reportable business segments does 2U, Inc. have?", A:``2U, Inc. has two reportable business segments."\}

\qquad  ...
\newline
\newline
\textbf{Financial Statement}: \{\textcolor{red}{Context}\}
\end{tcolorbox}
\caption{The prompt for generating QA pairs.}
\label{fig:an_loc}
\end{figure}

\begin{figure}[h]
\begin{tcolorbox}
[title=Example prompt used for generating \textbf{\textit{Reasoning}} QAs in financial statements]

\textbf{Task:} You are a financial analyst. Design a question requiring mathematical/logical derivation from the financial statement, adhering to:
\newline
\newline
\textbf{Design Requirements:}
\newline
\newline
1. Require multi-step processing of:

\qquad - Numerical calculations (e.g., ratios, growth rates)

\qquad - Temporal comparisons

\qquad - Reasonable inferences
\newline
\newline
2. Answerability Constraints:

\qquad - Unanswerable without the provided context

\qquad - Impossible to answer through simple lookup
\newline
\newline
3. Response Specifications:

\qquad - Maximum answer length: avoid too long

\qquad - Forbid speculative or probabilistic responses
\newline
\newline
\textbf{Example QAs}:

\qquad  \{Q: ``Calculate the percentage increase in total assets from December 31, 2023, to March 31, 2024, and explain what this increase indicates about the company's financial position during this period.", A:``Total assets increased from \$22,008,739 on December 31, 2023, to \$36,852,097 on March 31, 2024, which is an increase of \$14,843,358. The percentage increase is (14,843,358 / 22,008,739) * 100, which equals approximately 67.5\%. This significant increase in total assets indicates improved financial strength, potentially from additional capital inflows or asset acquisitions during the period."\}

\qquad  ...
\newline
\newline
\textbf{Financial Statement}: \{\textcolor{red}{Context}\}
\end{tcolorbox}
\caption{The prompt for generating QA pairs.}
\label{fig:an_rea}
\end{figure}

\begin{figure}[h]
\begin{tcolorbox}
[title=Example prompt used for generating \textbf{\textit{Comparison}} QAs in financial statements]

\textbf{Task:} You are a financial analyst. Create a comparative question requiring synthesis of two distinct sections from the financial statement, following these principles:
\newline
\newline
\textbf{Design Requirements:}
\newline
\newline
1. Must reference:

\qquad - Disparate metrics (e.g., departmental budgets vs regional sales)

\qquad - Chronological differences (quarterly/annual comparisons)

\qquad - Contrasting categories (actual vs projected figures)
\newline
\newline
2. Dependency Rules:

\qquad - Each segment provides unique essential information

\qquad - No overlapping data between required sections
\newline
\newline
3. Answerability Constraints:

\qquad - Unanswerable without the provided context

\qquad - Answer must demonstrate relational understanding

\qquad - Require explicit mention of both referenced sections
\newline
\newline
4. Response Specifications:

\qquad - Maximum answer length: avoid too long

\qquad - Comparisons must use contextually appropriate units
\newline
\newline
\textbf{Example QAs}:

\qquad  \{Q: ``How did the average revenue per Full Course Equivalent (FCE) enrollment change in the Degree Program Segment compared to the Alternative Credential Segment from 2022 to 2023?", A:``The average revenue per FCE enrollment increased by 17.8\% in the Degree Program Segment from \$2,447 in 2022 to \$2,883 in 2023, while it decreased by 6\% in the Alternative Credential Segment from \$3,897 in 2022 to \$3,662 in 2023."\}

\qquad  ...
\newline
\newline
\textbf{Financial Statement}: \{\textcolor{red}{Context}\}
\end{tcolorbox}
\caption{The prompt for generating QA pairs.}
\label{fig:an_comp}
\end{figure}

\begin{figure}[h]
\begin{tcolorbox}
[title=Example prompt used for generating \textbf{\textit{Hallucination Detection}} QAs in financial statements]

\textbf{Task:} You are a financial analyst. Design a pseudo-relevant question that appears answerable but actually lacks sufficient basis in the financial statement, following these principles:
\newline
\newline
\textbf{Design Requirements:}
\newline
\newline
1. Surface-level Relevance:

\qquad - Use document-specific terminology

\qquad - Reference actual sections/metrics as distractors
\newline
\newline
2. Unanswerability Guarantees:

\qquad - Absolutely not mentioned in the context

\qquad - Missing critical data points required for resolution

\qquad - No inferential path from provided information
\newline
\newline
3. Confirm absence of: 

\qquad - Direct mentions

\qquad - Implied values

\qquad - Comparable proxies
\newline
\newline
\textbf{Example QAs}:

\qquad  \{Q: ``What measures are being taken to mitigate foreign currency risk?", A:``The document does not specify any measures being taken to mitigate foreign currency risk."\}

\qquad  \{Q: ``What is the company's market share in the non-combustible nicotine-related products sector?", A:``The document does not provide a specific percentage for the company's market share in the non-combustible nicotine-related products sector."\}

\qquad  ...
\newline
\newline
\textbf{Financial Statement}: \{\textcolor{red}{Context}\}
\end{tcolorbox}
\caption{The prompt for generating QA pairs.}
\label{fig:an_hallu}
\end{figure}

\section{Task Examples}
For each task type, we provide three test examples, one for each type of context, to specifically demonstrate queries in LaRA that are close to real-world scenarios (Figure~\ref{fig:location},~\ref{fig:reasoning},~\ref{fig:comp}, and~\ref{fig:hallu}). Although we do not explicitly define lower-level sub-tasks within the four main task categories, it is evident from the examples that the emphasis of the same task varies across different context types. 
For example, financial statements are well-suited for mathematical calculations, while research papers are ideal for comparing different viewpoints and data.

We believe the most essential principle that all questions must adhere to is practicality—that is, the questions should reflect those that a person could realistically ask based on the given context. While questions designed solely to challenge LLMs have academic value, they do not address the considerations necessary for designing a real-world RAG or LC planning system. \emph{Our work aims to explore how to make more optimal choices under a realistic query distribution.}
\begin{figure}[h]
\small
\begin{tcolorbox}
[title=Examples of \emph{location} task]
\textbf{Novel} 
\newline
\newline
\textbf{[Query]} What does Mrs. Fitzgerald think of Violet and her family?
\newline
\textbf{[Ground-truth Answer]} She believes they are very common.
\newline
\newline
============================================================================

\textbf{Academic Paper} 
\newline
\newline
\textbf{[Query]} According to paper 2, what impact does self-critiquing have on the plan generation performance of LLMs in comparison to using an external verifier?
\newline
\textbf{[Ground-truth Answer]} Self-critiquing degrades the plan generation performance compared to using an external, sound verifier.
\newline
\newline
============================================================================

\textbf{Financial Statements} 
\newline
\newline
\textbf{[Query]} How much cash was provided by financing activities during the three months ended March 31, 2024?
\newline
\textbf{[Ground-truth Answer]} \$6.3 million.
\newline
\newline
\end{tcolorbox}
\caption{The examples of location task.}
\label{fig:location}
\end{figure}

\begin{figure}[h]
\small
\begin{tcolorbox}
[title=Examples of \emph{reasoning} task]
\textbf{Novel} 
\newline
\newline
\textbf{[Query]} Why did Richard know by the second meeting that Karen didn't like sweets?
\newline
\textbf{[Ground-truth Answer]} Because during their first meeting over a meal, Richard noticed that Karen quietly disposed of the dessert on her plate.
\newline
\newline
============================================================================

\textbf{Academic Paper} 
\newline
\newline
\textbf{[Query]} Based on paper 2, why does the LLM+LLM backprompting system underperform compared to the LLM+VAL system in plan generation?
\newline
\textbf{[Ground-truth Answer]} The LLM+LLM backprompting system underperforms because the verifier LLM produces a significant number of false positives, declaring incorrect plans as valid, which undermines the reliability of the system. In contrast, the LLM+VAL system uses an external sound verifier, VAL, which provides accurate validation, leading to better overall performance due to fewer verification errors.
\newline
\newline
============================================================================

\textbf{Financial Statements} 
\newline
\newline
\textbf{[Query]} What percentage of the total fair value of the liabilities assumed during the merger with DHC were warrant liabilities?
\newline
\textbf{[Ground-truth Answer]} To calculate the percentage of warrant liabilities in the merger with DHC, sum the warrant liability\'s value, \$1,913,737, and divide it by the total net liabilities assumed, \$9,863,196. The percentage is found by (1,913,737 / 9,863,196) * 100, which equals approximately 19.4\%.
\newline
\newline
\end{tcolorbox}
\caption{The examples of reasoning task.}
\label{fig:reasoning}
\end{figure}

\begin{figure}[h]
\small
\begin{tcolorbox}
[title=Examples of \emph{comparison} task]
\textbf{Novel} 
\newline
\newline
\textbf{[Query]} How does the narrator's attitude towards Turkey and Nippers compare to his attitude towards Bartleby?
\newline
\textbf{[Ground-truth Answer]} The narrator is willing to overlook Turkey's and Nippers' flaws due to their usefulness, just as he tolerates Bartleby's peculiar behavior because of his steady work and presence.
\newline
\newline
============================================================================

\textbf{Academic Paper} 
\newline
\newline
\textbf{[Query]} How do the approaches for incorporating context into ranking functions differ between the method proposed in paper 0 and in paper 1?
\newline
\textbf{[Ground-truth Answer]} Paper 0 incorporates context using delta features comparing neighboring items, while paper 1 uses a self-attention mechanism to account for interactions between items during both training and inference.
\newline
\newline
============================================================================

\textbf{Financial Statements} 
\newline
\newline
\textbf{[Query]} How did the total current assets and total current liabilities change for CISO Global, Inc. from December 31, 2023, to March 31, 2024?
\newline
\textbf{[Ground-truth Answer]} CISO Global's total current assets decreased from \$10,957,814 on December 31, 2023, to \$9,276,063 on March 31, 2024, while total current liabilities increased from \$26,071,102 to \$32,604,126 in the same period.
\newline
\newline
\end{tcolorbox}
\caption{The examples of comparison task.}
\label{fig:comp}
\end{figure}

\begin{figure}[h]
\small
\begin{tcolorbox}
[title=Examples of \emph{hallucination detection} task]
\textbf{Novel} 
\newline
\newline
\textbf{[Query]} What type of flower did Alexander place on Violet's grave in the cemetery?
\newline
\textbf{[Ground-truth Answer]} TThe text does not mention Alexander placing any type of flower on Violet's grave in the cemetery.
\newline
\newline
============================================================================

\textbf{Academic Paper} 
\newline
\newline
\textbf{[Query]} In paper 2, what are the implications of AI-enhanced NMR processing on the prediction of chemical reaction pathways?
\newline
\textbf{[Ground-truth Answer]} Paper 2 does not discuss the implications of AI-enhanced NMR processing on the prediction of chemical reaction pathways.
\newline
\newline
============================================================================

\textbf{Financial Statements} 
\newline
\newline
\textbf{[Query]} What were the results of the company's environmental sustainability initiatives?
\newline
\textbf{[Ground-truth Answer]} The financial statement does not mention or provide any details about the results of the company's environmental sustainability initiatives.
\newline
\newline
\end{tcolorbox}
\caption{The examples of hallucination detection task.}
\label{fig:hallu}
\end{figure}

\label{apx:case}
\end{document}